\documentclass[runningheads]{llncs}
\usepackage{graphicx}
\usepackage{verbatim}
\usepackage[english]{babel}
\usepackage[numbers]{natbib}
\usepackage{tabto}
\graphicspath{ {./media/} }
\usepackage{epstopdf}
\usepackage{caption}
\usepackage{array}
\usepackage{soul}
\newcolumntype{P}[1]{>{\centering\arraybackslash}p{#1}}

\begin{document}
\title{Glioma Segmentation with Cascaded Unet}
%
%
\author{Dmitry Lachinov\inst{1,2}\orcidID{0000-0002-2880-2887} \and
Evgeny Vasiliev\inst{1}\orcidID{0000-0002-7949-1919}
\and
Vadim Turlapov\inst{1}\orcidID{0000-0001-8484-0565}}
\authorrunning{D. Lachinov et al.}
%
\institute{Lobachevsky State University, Gagarina ave. 23, 603950 Nizhny Novgorod, Russian Federation 
\email{\\ \{dlachinov, eugene.unn, vadim.turlapov\}@gmail.com }
\and
Intel, Nizhny Novgorod, Russian Federation
\email{dmitry.lachinov@intel.com}}
\maketitle              
\begin{abstract}
MRI analysis takes central position in brain tumor diagnosis and treatment, thus it's precise evaluation is crucially important. However, it's 3D nature imposes several challenges, so the analysis is often performed on 2D projections that reduces the complexity, but increases bias. On the other hand, time consuming 3D evaluation, like, segmentation, is able to provide precise estimation of a number of valuable spatial characteristics, giving us understanding about the course of the disease.\newline 
Recent studies, focusing on the segmentation task, report superior performance of Deep Learning methods compared to classical computer vision algorithms. But still, it remains a challenging problem. In this paper we present deep cascaded approach for automatic brain tumor segmentation. Similar to recent methods for object detection, our implementation is based on neural networks; we propose modifications to the 3D UNet architecture and augmentation strategy to efficiently handle multimodal MRI input, besides this we introduce approach to enhance segmentation quality with context obtained from models of the same topology operating on downscaled data. We evaluate presented approach on BraTS 2018 dataset and discuss results.

\keywords{segmentation \and BraTS \and UNet \and cascaded UNet \and deep learning}
\end{abstract}

\section{Introduction}
Multimodal magnetic resonance imaging (MRI) is a powerful tool for studying human brain. Among it's different applications, it is mainly used for disease diagnosis and treatment planning. Accurate assessment of MRI results is critical throughout all these steps.
Since MRI scans are the set of multiple three dimensional arrays, it's manual analysis and evaluation is a non-trivial procedure and requires time, attention and expertise. Lack of these resources can lead to unsatisfying results. Typically, these scans are analyzed by clinical experts using two dimensional cut and projection planes. It limits the amount of data taken into account for decision making, this it adds bias to the resulting evaluation. On the other hand, accurate segmentation and 3D reconstruction is able to provide more insights on disease progression and help a therapist to plan the treatment better. However these methods are not widely used due to unreasonable amount of time needed for manual labeling.\newline
Denoting the problem of automatic glioma segmentation Brain Tumor Segmentation (BraTS) challenge \cite{dataset2, dataset1} was created and became an annual competition allowing participants to evaluate and compare their state of the art methods using unified framework. Participants are called to develop their algorithms and produce segmentation labels of the different glioma sub-regions: "enhancing tumor" (ET), "tumor core" (TC) and "whole tumor" (WT). The training data \cite{dataset3,dataset4} consists of 210 high grade and 75 low grade glioma MRIs manually labeled by experts in the field. Testing data is split into two parts: \textbf{validation set} that can be used for evaluation throughout the challenge and \textbf{test set} for final evaluation. Performance of the methods is measured using Dice coefficient, Sensitivity, Specificity and Hausdorff distance.\newline
Above-named challenge made a significant impact on the evolution of computational approaches for tumor segmentation. In the last few years, a variety of algorithms were proposed to solve this problem. Compared with other methods, convolutional neural networks have been showing the best state of the art performance for computer vision tasks in general and for biomedical image processing tasks in particular.\newline
In this paper we present cascaded variant of the popular UNet network \cite{unet,3dunet} that iteratively refines segmentation results of it's previous stages. We employ this approach for brain tumor segmentation task in the scope of BRATS 2018 challenge and evaluate it's performance. We also compare regular 3D UNet \cite{3dunet} with it's cascaded counterpart.\newline
\section{Method}
\begin{figure}[ht]
\includegraphics[scale=1.8]{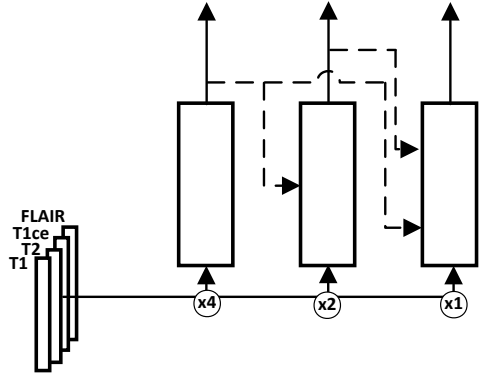}
\centering
\captionsetup{justification=centering}
\caption{Schematic representation of approach employed in this paper. T1, T2, T1ce, FLAIR stands for input MRI modalities. x4, x2 indicate downsampling factor for the network input. Dotted arrows indicate connections between networks $C_i$ that are illustrated as basic blocks.}
\label{fig:pipeline}
\end{figure}

In this study we propose neural networks based approach for brain tumor segmentation. Our method can be represented as a chain of multiple classifiers $C_i$ of the same topology $F$ refining segmentation output of previous iterations. Every classifier $C_i$ shares the same topology but has it's own set of parameters $W_{i}$ that is subject to optimization during training. $Y_{i}$ - the result of the i-th step can be represented as
$Y_{i} = F(X_{i},Y_{i-1},Y_{i-2},W_{i})$, where $X_{i}$ is the i-th input.\newline
Described approach is illustrated in figure \ref{fig:pipeline}. Each of the basic blocks $C_{i}$ is a UNet network modified with respect to the task of glioma segmentation. Compared to the original UNet architecture described in \cite{unet} and extended for 3D case in \cite{3dunet}, we employ multiple encoders separately handling input modalities and introduce the way to merge their output.
\newline

\subsection{Multiple encoders UNet}
\begin{figure}[ht]
\includegraphics[scale=1.0]{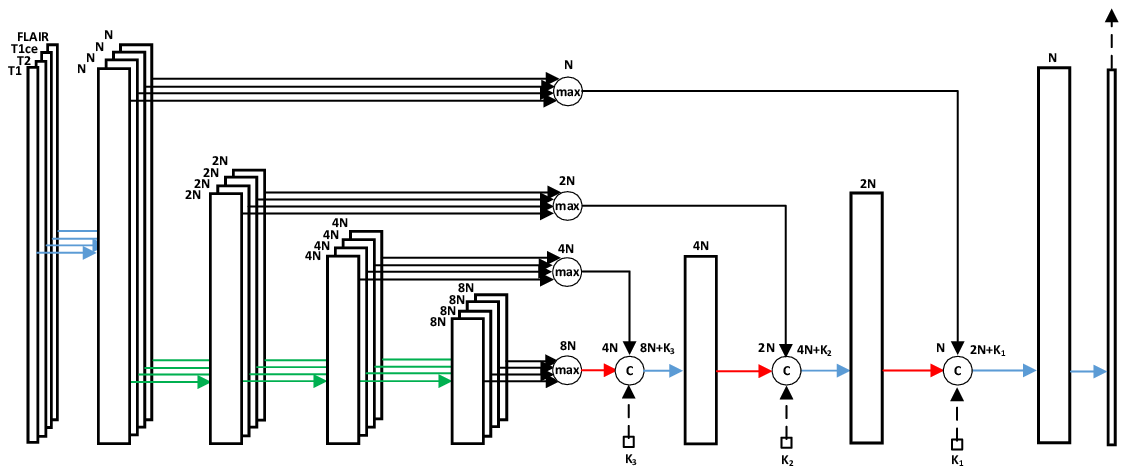}
\includegraphics[scale=1.2]{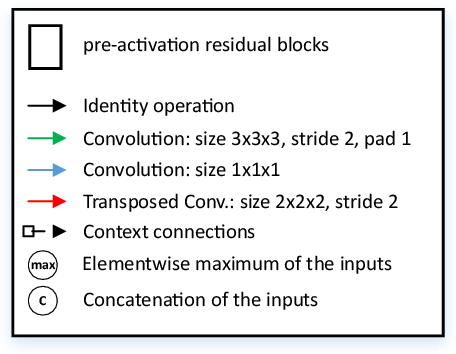}
\centering
\captionsetup{justification=centering}
\caption{Architecture of multiple encoders UNet. T1, T2, T1CE, FLAIR stand for input modalities. $N$ is a base number of filters, $K$ is a number of filters in context feature map obtained from lower scale models.}
\label{fig:unet}
\end{figure}

\begin{figure}[ht]
\includegraphics[scale=1.2]{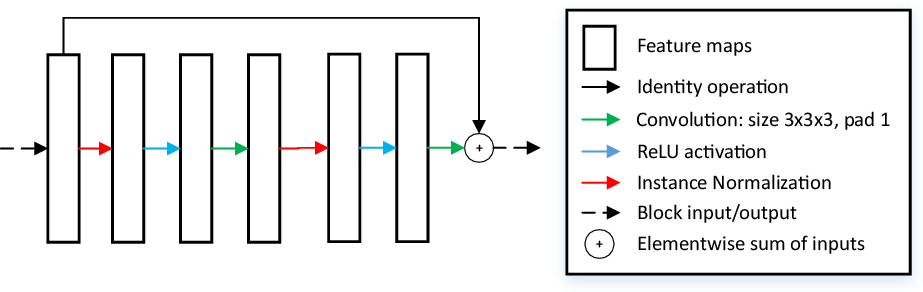}
\centering
\captionsetup{justification=centering}
\caption{Design of the residual block}
\label{fig:resblock}
\end{figure}

Traditional UNet architecture \cite{unet} extended for handling volumetric input \cite{3dunet} has two stages: encoder part where network learns feature representations on different scales and aggregates contextual information, and decoder part where network extracts information from observed context and previously learned features. Skip connections employed between corresponding encoder and decoder layers enable efficient training of the deep parts of the network and comparison of identically scaled features with different receptive fields.\newline
This method allows to handle multimodal MRI input, however, it mixes and processes signals of different types identically. In contrast, our approach learns feature representations for every modality separately and combines them at later stages. This is achieved by employing grouped convolutions in the encoder path with number of groups equals to the number of input modalities. Resulting features are calculated as a maximum of the feature maps produced by encoders. In order to preserve feature maps' sizes we employ point-wise convolution right after max operation. Similar to the original UNet, the number of filters is doubled with every downsampling operation and reduced by half with every upsampling operation, ReLU is used as activation function after every convolution layer. Described architecture is illustrated in figure \ref{fig:unet}.\newline
The network is built of basic pre-activation residual blocks \cite{resnet} that consist of two instance normalization layers, two relu activation layers and two convolutions with kernel size 3. This basic building block is illustrated in figure \ref{fig:resblock}.

\subsubsection{Cascaded UNet}
Proposed network is illustrated in figure \ref{fig:pipeline} and consists of three basic blocks. Each block by itself is a modified UNet network with it's own loss function at the end. Every next block takes downsampled volume as an input and produces segmentation of the corresponding size. Similar to DeepMedic \cite{deepmedic}, this architecture simultaneously processes the input image at multiple scales and extracts scale-specific features. The feature map before the last convolutional layer in every block is concatenated to the corresponding feature map of higher-scale block. It enables the context information flow between networks with different scales.\newline
In UNet architecture decoder output at each scale $i$ depends on encoder output at the same scale (skip connections) and decoder output of the previous scale: $d^{t}_{i}=f(e^{t}_{i},d^{t}_{i-1})$, where $d^{t}_{i}$ is decoder output, and $e^{t}_{i}$ is encoder output at scale $i$, and t is the index of the network. Expanding the first convolution of $f$ we get $d^{t}_{i}=g(W^{t}_{i,e}e^{t}_{i}+W^{t}_{i,d}d^{t}_{i-1})$, where $W$ are trainable parameters. Here we propose to incorporate context of the lower scale networks by concatenating corresponding network output $y^t$ (see figure \ref{fig:unet}, illustrated as dotted arrows) so $d^{t}_{i}$ becomes $d^{t}_{i}=g(W^{t}_{i,e}e^{t}_{i}+W^{t}_{i,d}d^{t}_{i-1}+W^{t}_{i,y}y^{t-i})$. This approach fuses multiple networks operating at different scales together and encourages model to iteratevily refine results of previous iterations.\newline

\subsection{Preprocessing \& Data augmentation}
\begin{figure}[ht]
\includegraphics[scale=0.25]{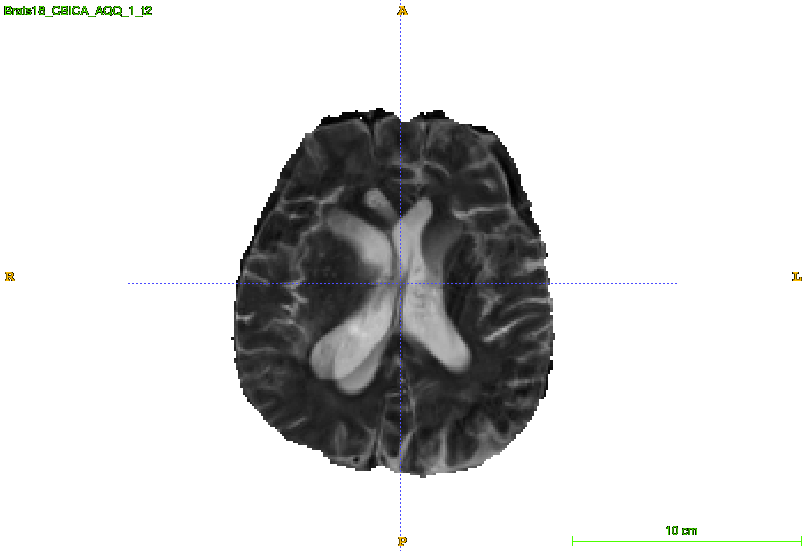}
\centering
\captionsetup{justification=centering}
\caption{Example of the registration artifacts found in the training dataset. Visualization is done with ITK-SNAP \cite{itksnap}.}
\label{fig:artifacts}
\end{figure}

We have found data preprocessing employed in \cite{norm} to be especially effective. Like in \cite{norm}, we perform z-score normalization on non-zero (brain) voxels. After that we are eliminating outliers and noise by clamping all values to the range from -5 to 5. At the final step we shift brain voxels to the range [0;10] and assign zeros to background.\newline
For offline data augmentation we artificially increase number of samples by employing b-spline transformation to the original data. It has been done with ITK implementation \cite{itk}.\newline
During training we randomly flip input image along sagittal plane and "mute" input modalities with predefined probability. Without this augmentation the network was only considering one of the input modalities while making a prediction and not taking others into account. To deal with this issue we are randomly filling input channels with Gaussian noise. We introduce probability to apply this augmentation for every channel and set it to 0.1, so there is 34\% chance to mute at least one out of four modalities. This also helps to aggregate information allover input data and to deal with noisy or corrupted input images like illustrated in the picture \ref{fig:artifacts}.\newline

\subsection{Training}
The training procedure is conducted on brain regions resampled to 128x128x128 voxels. We are operating with downsampled data to preserve the context since we believe it plays important role for robust segmentation of multimodal MRI scans obtained from different institutions and scanners. We use Mean Dice loss $L_{mean\_dice}(g,p)$ where $g$ is a ground truth, $p$ is a model's prediction. We trained our network with stochastic gradient descent with initial learning rate of 0.1, exponential learning rate decay with rate 0.99 for every epoch, weight decay of 0.9 and minibatch size equal to 4 samples.

$$L_{mean\_dice}(g,p) = 1 - \frac{1}{|C|} \sum_{c\in C}{ \frac{\sum_{i_c}{p_{c}^{i}g_{c}^{i}}}{\sum_{i_c}{p_{c}^{i}+g_{c}^{i}}}},$$ where $C$ is a set of different classes.\newline
This CNN was implemented in MXNet framework \cite{mxnet} and trained using four GTX 1080TI with batch size 4 to enable data parallelism. Training was performed for 500 epoches.

\section{Results}
In this section we report evaluation results obtained with online validation system provided by organizers. With intention to penalize model for relying on the one single modality we apply channel-out augmentation to the input data by randomly filling input modalities with Gaussian noise in addition to standard augmentations like mirroring and elastic transformations. Then we compare results obtained with this augmentation disabled (table \ref{tab2}) and enabled (table \ref{tab3}). The challenge validation data \cite{dataset3,dataset4} contains 66 MRI scans obtained with different scanners and from different institutions. Results of evaluation on validation dataset are reported in table \ref{tab4}.\newline

\begin{table}
\vspace{-4mm}
\caption{Evaluation of glioma segmentation without channel-out augmentation; Dice index is reported, WT stands for whole tumor, ET stands for enhancing tumor and TC stands for tumor core.}
\label{tab2}
\centering
\begin{tabular}{|P{2.0cm}|P{1.0cm}|P{1.0cm}|P{1.0cm}|}
\hline
Method & WT & ET & TC\\
\hline
UNet & 0.901 & 0.767 & 0.797 \\
ME UNet & 0.904 & 0.763 & 0.823\\
C ME UNet & 0.906 & 0.772 & 0.836\\
\hline
\end{tabular}
\vspace{-8mm}
\end{table}

\begin{table}
\vspace{-4mm}
\caption{Evaluation of glioma segmentation with channel-out augmentation; Dice index is reported, WT stands for whole tumor, ET stands for enhancing tumor and TC stands for tumor core.}
\label{tab3}
\centering
\begin{tabular}{|P{2.0cm}|P{1.0cm}|P{1.0cm}|P{1.0cm}|}
\hline
Method & WT & ET & TC\\
\hline
UNet & 0.901 & 0.779 & 0.837 \\
ME UNet & 0.907 & 0.784 & 0.827\\
C ME UNet & 0.908 & 0.784 & 0.844\\
\hline
\end{tabular}
\vspace{-8mm}
\end{table}

\begin{table}
\caption{Performance of proposed method on BraTS 2018 validation data, Dice index is reported.}
\label{tab4}
\centering
\begin{tabular}{|P{2.0cm}|P{1.0cm}|P{1.0cm}|P{1.0cm}|}
\hline
 & WT & ET & TC\\
\hline
Mean & 0.908 & 0.784 & 0.844 \\
StdDev & 0.065 & 0.237 & 0.161\\
Median & 0.926 & 0.858 & 0.906\\
25quantile & 0.9 & 0.805 & 0.791\\
75quantile & 0.943 & 0.897 & 0.947\\
\hline
\end{tabular}
\end{table}

\section{Conclusion}
In this paper we presented automatic segmentation algorithm solving two main problem arising during brain tumor segmentation with multimodal scans: complex input consisting of multiple modalities and overconfidence of the classifier. Solving the problem of heterogeneous input we proposed to use multiple encoders, so that every individual input modality produces corresponding feature maps independently from others; and we introduced the way to merge encoded feature maps. Also we explored influence of channel-out augmentation on model's output quality and we showed that proposed architecture benefits from this aggressive augmentation. It encourages model to take into account whole input by implicitly penalizing classifiers that rely only on one single modality. As a result model becomes robust to the presence of noise and corrupted data that could be encountered in the training and validation datasets. Moreover we introduced the way to efficiently fuse multiple models operating on the different resolution that forms a cascade of classifiers. Every next classifiers takes results of previous ones and refines the segmentation for it's specific scale. It enables iterative result refinement with less parameters than in corresponding deep models.
As a part of BraTS 2018 challenge \cite{dataset1,dataset2} we implemented and evaluated our approach with online validation tools. As a result we achieved high mean score and notably high median score. The mean Dice score of 0.908/0.784/0.844 was reported on validation dataset for the Whole tumor, Enhancing tumor and Tumor core correspondingly \newline
\newpage

\begingroup
\let\clearpage\relax
\bibliographystyle{splncs04}
\bibliography{samplepaper}
\endgroup
\end{document}